# ThreshNet: An Efficient DenseNet Using Threshold Mechanism to Reduce Connections

Rui-Yang Ju[1], Ting-Yu Lin[2], Jia-Hao Jian[1], Jen-Shiun Chiang[1], and Wei-Bin Yang[1]
[1]Department of Electrical and Computer Engineering, Tamkang University, New Taipei City 25137, Taiwan
[2]Department of Engineering Science, National Cheng Kung University, Tainan City 70101, Taiwan
https://github.com/RuiyangJu/ThreshNet

*Abstract*—With the continuous development of neural networks for computer vision tasks, more and more network architectures have achieved outstanding success. As one of the most advanced neural network architectures, DenseNet shortcuts all feature maps to solve the model depth problem. Although this network architecture has excellent accuracy with low parameters, it requires an excessive inference time. To solve this problem, HarDNet reduces the connections between the feature maps, making the remaining connections resemble harmonic waves. However, this compression method may result in a decrease in the model accuracy and an increase in the parameters and model size. This network architecture may reduce the memory access time, but its overall performance can still be improved. Therefore, we propose a new network architecture, ThreshNet, using a threshold mechanism to further optimize the connection method. Different numbers of connections for different convolution layers are discarded to accelerate the inference of the network. The proposed network has been evaluated with image classification using CIFAR 10 and SVHN datasets under platforms of NVIDIA RTX 3050 and Raspberry Pi 4. The experimental results show that, compared with HarDNet68, GhostNet, MobileNetV2, ShuffleNet, and EfficientNet, the inference time of the proposed ThreshNet79 is 5%, 9%, 10%, 18%, and 20% faster, respectively. The number of parameters of ThreshNet95 is 55% less than that of HarDNet85. The new model compression and model acceleration methods can speed up the inference time, enabling network models to operate on mobile devices.

## I. INTRODUCTION

In recent years, neural networks have reached high achievements in the field of computer vision (CV); an increasing number of network models are being widely used in various related tasks. With the continuous improvement of computing power in mobile devices, it is possible to perform image recognitions on a mobile computing platform. Because embedded platforms are limited in terms of computing power, storage capacity, and battery power, lightweight networks have to meet the conditions of a small model size, low power consumption, and low computational complexity.

Techniques for model compression and acceleration are hot research topics in the field of neural networks. They cannot only improve model performance, but also speed up the inference response and reduce server resource usages. Model compression usually focuses on reducing the number of parameters of the network model, while model acceleration focuses on reducing computational complexities and improving parallelism. Model compression and model acceleration can be studied in terms of algorithms, frameworks, and hardware. The algorithm directions mainly include structural optimization, model pruning, and model distillation.

GoogLeNet [1] uses global average pooling to replace the fully connected layer. Because the fully connected layer accounts for approximately 90% of the parameters, this approach can significantly reduce the number of parameters. Group convolution is widely used in convolutional neural network models, such as ShuffleNet V1, V2 [2][3] and MobileNet V1, V2 [4][5]. In order to reduce the numbers of parameters of the model, these networks first divide the feature maps into different groups and then fuse the results after their respective processing. The shared weight can greatly reduce the number of parameters by sharing the same parameters. For example, ALBERT [6] reduces the original number of parameters to $1/12$. However, because the amount of computation cannot be reduced, the inference time is not accelerated. Factorized convolution substitutes small convolution kernels for large convolution kernels to reduce the number of parameters. Inception V2 [7] replaces a $5 \times 5$ convolution kernel with two $3 \times 3$ convolution kernels, while Inception V3 [8] splits a $7 \times 7$ convolution kernel into a $1 \times 7$ convolution kernel and a $7 \times 1$ convolution kernel, and therefore the number of parameters is greatly reduced with the same convolution effect. SqueezeNet [9] replaces the $3 \times 3$ convolution with a $1 \times 1$ convolution and reduces the number of channels in the $3 \times 3$ convolution to increase the operation speed. GhostNet [10] uses a combination of linear kernels and convolution kernels to form many ghost modules to reduce the amount of computation. EfficientNet V1, V2[11][12] use compound coefficients to uniformly scale the network depth, width, and image resolution, which reduce the number of parameters while ensuring accuracy.

Model pruning [13] can be categorized as synaptic pruning, neuron pruning, and weight matrix pruning. Synaptic pruning is a method to remove unimportant connections between neurons, which involves the direct removal of a node. Weight

Corresponding Author: Jen-Shiun Chiang (e-mail: chiang@mail.tku.edu.tw)

matrix pruning sets unimportant parameters to 0. Michel [14] analyzed the role of each head in the multi-head mechanism in BERT [15], and analyzed it by removing the heads and adding masks. The results show that it is advisable to directly remove the entire weight matrix, which greatly improves the compression rate compared with the above pruning methods.

Model distillation [16][17] is to fit the new model with the original model. For example, DistillBERT [18] removes one layer every two layers, and the second layer of the new model corresponds to the third layer of the original model. The speed of the model is 60% faster but 60% smaller than that of BERT [15]. TinyBERT [19] fits the embedding layer, transformer layer, and prediction layer, reducing the model size by 7.5 times while speeding up the inference time by 9.4 times.

This paper proposes a model acceleration method that is different from the above researches. According to the newly proposed rules of the threshold mechanism, the connections between layers can be sparsed to improve the speed of the inference response. In this work, ThreshNet, was inspired by the concept of threshold voltage of a MOSFET, and we apply the concept of threshold voltage conversion to the layer-to-layer shortcut connections. The threshold mechanism is used for the first time as a tool to judge different statuses in the network. Similar to the threshold voltage in a MOSFET, we regard the different number of layers in a block as the value of voltage, and connect the feature maps of different layers in different methods according to the number of layers. The different methods are improved based on existing advanced network architectures. The proposed methodology can reduce the number of parameters of the network and the amount of model computation to minimize the inference time. Therefore, ThreshNet prunes the dense connections of DenseNet [20], uses the idea of harmonic dense connections from HarDNet [21], and further improves it. The proposed network architecture is better than HarDNet in terms of the accuracy, model size, and inference time.

There are two major contributions in the proposed approach.

1) This article proposes a new model compression and acceleration method. According to the threshold mechanism, the connection between layers is sparsed to different degrees, reducing the amount of calculation and improving the inference speed.

2) ThreshNet can overcome the side effect of increasing the number of parameters due to the increased layer weights of HarDNet. This proposed neural network architecture can further reduce the inference time and improve accuracy.

## II. RELATED WORKS

*A. Densely Connected Network*

DenseNet [20] adopts the concept of shortcut connections proposed by ResNet [22] to the extreme. It connects each layer with all the previous layers in the channel dimension and uses it as the input of the next layer. For an *L*-layer network, DenseNet eventually obtains $[L(L + 1)]/2$ connections, and directly splices the feature maps of all different layers to achieve feature reuse and improves the efficiency. Liao *et al.* [23] illustrates that ResNet is similar to the recurrent neural networks, but has more references than the recurrent neural networks. Because the features of ResNet cannot be reused in the layers, each layer needs to learn new weights, such that ResNet has a large number of parameters. Shortcuts can preserve most of the information, and therefore DenseNet performs global dense connections such that each layer can obtain the weight information of the previous layers through shortcuts.

In dense block, the feature maps of each layer have the same size and can be connected in the channel dimension. Inspired by He *et al.* [24], DenseNet defines $H(\cdot)$ as a composite function of three consecutive operations, batch normalization (BN) [7], rectified linear unit (ReLU) [25], and convolution (Conv). In addition, all layers of the dense block output $k$ feature maps after convolution operation. $k$ is called growth rate, which is a hyperparameter. Generally, a higher growth rate results in a larger output channel width. Owing to feature reuse, as the number of layers increases, the input of the next block will be very large even though $k$ is set small.

It has been shown in [8] that the amount of calculation can be reduced by adding a $1 \times 1$ convolution before the $3 \times 3$ convolution of the bottleneck layer. Since the input of the latter layers would be very large, a dense block can internally use bottleneck layers to reduce the computation. This is mainly to add a $1 \times 1$ convolution to the original architecture, where the $1 \times 1$ convolution obtains a $4k$ feature map, which plays a role in reducing the number of calculation.

*B. Harmonic Densely Connected Network*

The dense connection architecture of DenseNet causes a large amount of memory access time, resulting in excessive data transmission and considerable inference time. However, HarDNet [21] reduces the number of shortcuts between layers to reduce data traffic and speeds up the calculation time.

In HarDNet, Chao *et al*. used the concept of overlapping the harmonic waves to propose a new connection pruning method for DenseNet, and it is different from LogDenseNet [26]. In this process, if there exists a number *n*, where it is a non-negative integer and $k-2^n \geq 0$ and $k$ is divisible by $2^n$, a shortcut connects layer $k$ to layer $k-2^n$. Here the 0*th* layer is the input layer. By connecting in this way, once the $2^n th$ layer has been processed, it can discard most of the connections of the odd-numbered layers from layer 1 to layer $2^n$, and therefore it can reduce the memory access time significantly. This connection model resembles FractalNet [27], however they are different from those in which the latter uses average shortcuts instead of serial connections.

Since there exists an initial growth rate $k$ in the $Lth$ layer, HarDNet sets the number of channels to $k \times m^n$, where $n$ refers to the largest value of layer $L$ divisible by $2^n$, and multiplier $m$ is the dimension compression factor. If the input of the 0*th* layer has $k$ channels and $m = 2$, the channel ratio of each layer is 1 : 1. When *m* is set to be less than 2, it is equivalent to compressing the input channels into fewer output channels. In HarDNet, Chao *et al.* [21] argued that *m* is set to between 1.6 and 1.9 and it would obtain good accuracy and efficient parameters.

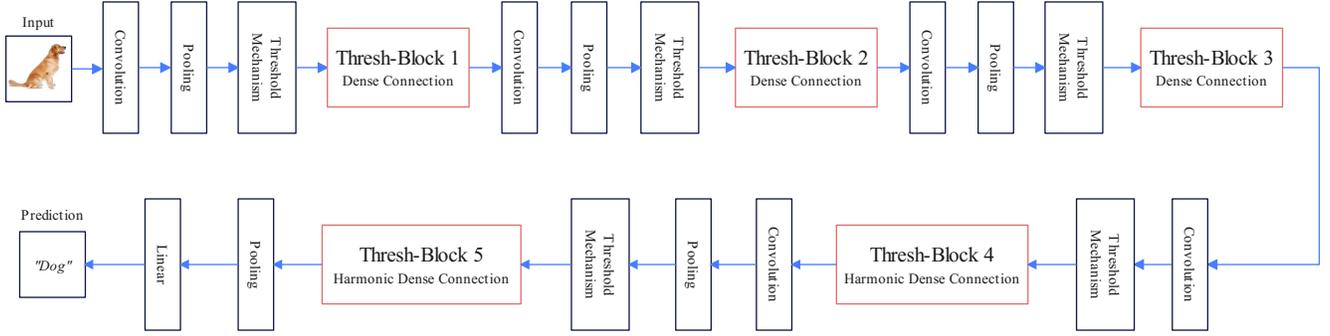

**FIGURE 1.** Example of ThreshNet79 model architecture. The first three Thresh-Blocks are densely connected, and the last two Thresh-Blocks are harmonically densely connected. The layer between two adjacent blocks is called the transition layer that changes the size of the feature map by convolution and pooling except for the third Thresh-Block.

Like DenseNet, a transition layer is added to HarDNet to reduce the size of the feature maps. For the transition layer, maxpooling is used for downsampling, and the nonlinear combination function $H(\cdot)$ uses a *Conv-BN-ReLU* structure.

## III. THRESHNET

### A. Threshold Mechanism

We propose the concept of the threshold mechanism for the network architecture, which is similar to the threshold voltage in MOSFET. In our proposed network architecture, the blocks are called Thresh-Blocks, and are differentiated by a certain value (threshold value) of the shortcut connection numbers in each layer.

Before entering each Thresh-Block, our system architecture examines the number of input channels and determine whether they exceed the threshold value. If the number exceeds the threshold value, the method for connecting the feature map of the layer would be changed. This procedure is called the threshold mechanism. In this work, the numbers of the input channels for all the Thresh-Blocks of the network architecture are set to 128, 192, 288, 480, and 960, respectively. Therefore, the number of the input channels of the Thresh-Block is set to 320 as the threshold value (according to Section IV.A., the median or average of the numbers of the input channels of all the Thresh-Blocks is selected as the threshold value). The number of the input channels in a Thresh-Block is related to the number of layers in the previous Thresh-Block. As shown in FIGURE 1, we consider the overall network architecture of ThreshNet79 as an example. The threshold mechanism performs condition judgments before Thresh-Blocks 1, 2, and 3. Since the numbers of the input channels of these three Thresh-Blocks do not exceed the threshold value, the mechanism of the connection changing methods is not executed. However, the number of the input channels in the fourth Thresh-Block has reached the threshold value, and the threshold mechanism would be activated.

Obviously, the numbers of input and output channels of the first several blocks are small; therefore, the sparse effect is insignificant. However, due to the large number of layers in the following blocks, if each layer is connected to the others, it would cause a significant amount of computation. Meanwhile the excessive number of connecting channels may require a large amount of memory access time. Under such circumstances, reducing the number of connections between layers can result in a good sparse effect and achieve good operational efficiencies.

Convolutional neural networks, such as GoogleNet [1], usually try to improve the accuracy of image classification results by deepening and widening the model. However, simply increasing the depth or width would aggravate gradient vanishing problems; the weight cannot be updated effectively, and finally the neural network cannot be trained further. DenseNet [20] uses global dense connections, which makes backpropagation easier, and the effect of the model convergence becomes much better. However, the computation of backpropagation is complex, and the memory usage during model training becomes dense. Different from the global dense connection of DenseNet, ThreshNet adds a threshold mechanism that uses different connection methods in different blocks to make backpropagation easier. Besides, it preferentially selects the features with fewer calculations for reuse, which can greatly reduce the inference time and reduce the memory usage for model training.

### B. Thresh Blocks

Comparing with the 4-block structure used in DenseNet, we expand the number of blocks to 5 while reducing the connection between the layers in the Thresh-Block. The proposed architecture can increase the model depth and effectively compress DenseNet to reduce the complexity of the model.

An important characteristic of the convolutional neural network is the downsampling layer, which changes the size of the feature maps. In the proposed network architecture, a $1 \times 1$ convolution layer and $2 \times 2$ average pooling layer are employed as the transition layer. Specifically, to ensure the size of the feature map, before Thresh-Block 4, where the threshold mechanism is activated, we only keep the $1 \times 1$ convolution layer. In general, Thresh-Block is composed of translation layer and convolution layer.

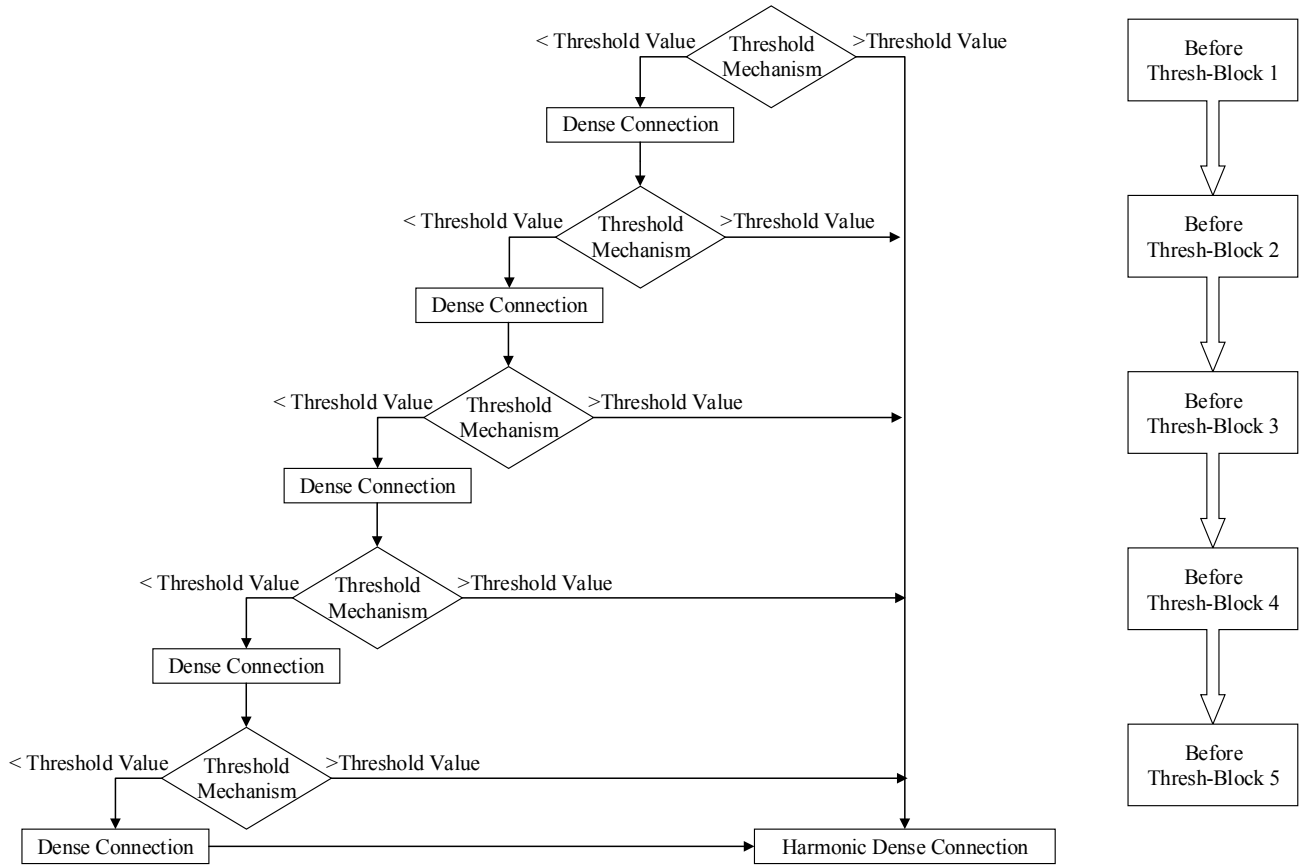

**FIGURE 2.** Use Threshold mechanism as a divider to distinguish between different connection method.

In the threshold mechanism, we have introduced the mechanism to judge the number of input channels of Thresh-Block. In Thresh-Block, the number of the input channels of each layer is multiplied by the growth rate to increase the number of feature maps. The growth rate details are described in the next subsection (III.C). As an entire Thresh-Block, its output channels are scaled down (the reduction rate is generally set to 0.5-0.85).

An excessive number of channels may affect the efficiency of model compression; therefore, counting the number of channels is the most straightforward method for determining the number of layers. FIGURE 2 shows the function of changing the connection methods of the threshold mechanism. Depending on the number of channels in the block, the network architecture uses dense connections in the first few Thresh-Blocks and harmonic dense connections in the rest of the Thresh-Blocks. For the characteristics of dense connection and harmonic dense connection in the network, the first three and last two Thresh-Blocks use different connection methods.

*C. Dense Connection*

**1) Connectivity**

For Thresh-Blocks, in which the threshold mechanism is judged not to trigger, the proposed network architecture uses dense connections from any layer to all the subsequent layers that are directly connected. FIGURE 3(a) shows the dense connection method, where the *lth* layer receives all the feature maps of the previous layers as inputs:

$$x_l = H_l([x_0, x_1, \cdots, x_{l-1}]) \quad (1)$$

where $[x_0, x_1, \cdots, x_{l-1}]$ refers to the concatenation of the feature maps generated in layers $0, 1, \cdots, l-1$.

**2) Composite Function**

In Equation (1), $H_l(\cdot)$ is a composite function of three consecutive operations: BN, ReLU, and $3 \times 3$ convolution.

**3) Growth Rate**

In ThreshNet, each function $H_l$ can generate $k$ feature maps, and the *lth* layer has $k_0 + k \times (l-1)$ input feature maps, where $k_0$ is the number of channels in the input layer. For the densely connected part of Thresh-Blocks, $k$ is set to 32, where $k$ is also called the growth rate of the network. More specifically, the number of input channels in each layer in Thresh-Blocks increases by 32, while the number of output channels is fixed at 32.

**4) Bottleneck Layer**

In order to ensure the accuracy of the model training, the dense connections for the Thresh-Blocks employ the bottleneck layer architecture. A $1 \times 1$ convolution is adopted as the bottleneck layer in front of each $3 \times 3$ convolution to reduce the number of input feature maps and thus improve the computation efficiency. The number of output channels of the 1 × 1 convolution is set to 4k. The input channel number of the subsequent 3 × 3 convolution is 4k, and the output channel number is k.

TABLE I
THRESHNET ARCHITECTURES

| Layers | Output Size | ThreshNet79 | ThresholdNet95 |
|---|---|---|---|
| Initial Conv | 112, 112 | 3 × 3, stride 2 | |
| | 112, 112 | 3 × 3, stride 1 | |
| Pooling | 56, 56 | 3 × 3, stride 2 | |
| Thresh-Dense Connection (1) | 56, 56 | $\begin{bmatrix}1\times 1\ \text{Conv}\\3\times 3\ \text{Conv}\end{bmatrix}\times 6$, k = 32 | $\begin{bmatrix}1\times 1\ \text{Conv}\\3\times 3\ \text{Conv}\end{bmatrix}\times 6$, k = 32 |
| Conv & Pooling (1) | 56, 56 | 1 × 1, stride 1 | |
| | 28, 28 | 2 × 2, stride 2 | |
| Thresh-Dense Connection (2) | 28, 28 | $\begin{bmatrix}1\times 1\ \text{Conv}\\3\times 3\ \text{Conv}\end{bmatrix}\times 8$, k = 32 | $\begin{bmatrix}1\times 1\ \text{Conv}\\3\times 3\ \text{Conv}\end{bmatrix}\times 12$, k = 32 |
| Conv & Pooling (2) | 28, 28 | 1 × 1, stride 1 | |
| | 14, 14 | 2 × 2, stride 2 | |
| Thresh-Dense Connection (3) | 14, 14 | $\begin{bmatrix}1\times 1\ \text{Conv}\\3\times 3\ \text{Conv}\end{bmatrix}\times 12$, k = 32 | $\begin{bmatrix}1\times 1\ \text{Conv}\\3\times 3\ \text{Conv}\end{bmatrix}\times 16$, k = 32 |
| Conv (3) | 14, 14 | 1 × 1, stride 1 | |
| Threshold Mechanism | 14, 14 | Threshold Mechanism | |
| Thresh-Harmonic Dense Connection (4) | 14, 14 | [3 × 3 Conv] × 16, k = 40 | [3 × 3 Conv] × 16, k = 40 |
| Conv & Pooling (4) | 14, 14 | 1 × 1, stride 1 | |
| | 7, 7 | 2 × 2, stride 2 | |
| Thresh-Harmonic Dense Connection (5) | 7, 7 | [3 × 3 Conv] × 4, k = 160 | [3 × 3 Conv] × 4, k = 160 |
| Classification | 1, 1 | AvgPool | |
| | | 1000D Fully-Connected | |

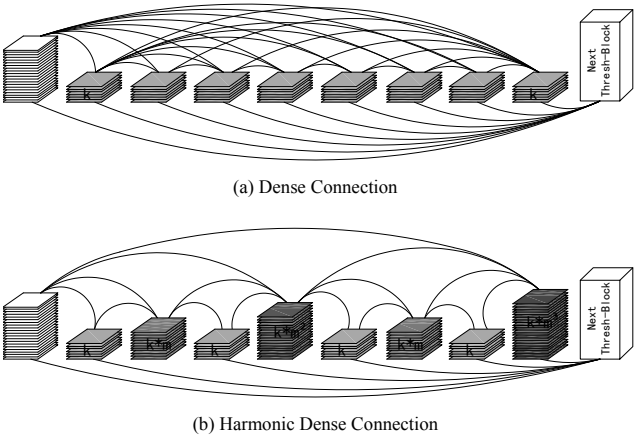

FIGURE 3. (a) Dense Connection, (b) Harmonic Dense Connection, and each cuboid represents a stack of multiple feature maps.

*D. Harmonic Dense Connection*

**1) Connectivity**

As shown in FIGURE 3 (b), if $2^n$ is divisible by $l$, the $lth$ layer would be connected to the $(l-2^n)th$ layer, where $n$ is a non-negative integer and $l - 2^n \geq 0$. The relationship between $x_l$ and $H_l$ is expressed as:

$$\begin{cases} if\ l\ \%\ 2 = 1, x_l = H_l([x_{l-1}]) \\ else, x_l = H_l([x_0,\cdots,x_{l-2^n}]) \end{cases} \qquad (2)$$

**2) Weighting**

In order to enable ThreshNet to operate on devices with less computing power, such as mobile phones, Raspberry Pi, and FPGA, we use the threshold mechanism to prune the global dense connections of DenseNet. However, it is not advisable to directly discard the connection between some layers for the block that uses the harmonic dense connection by the threshold mechanism, because it would reduce the accuracy of the model training results. Therefore, we reset the weights for the convolution layers in the Thresh-Block using harmonic dense connections and increase the number of channels for the convolution layers that retain connections to reduce the loss of feature extraction owing to discarding connections.

TABLE II
ABLATION STUDY

|  | ThreshNet 79-a | ThreshNet 79-b | ThreshNet 79-c | ThreshNet 79-d |
|---|---|---|---|---|
| Dense Connection | 6 | 6, 8 | 6, 8, 12 | 6, 8, 12, 16 |
| Harmonic Dense Connection | 8, 12, 16, 4 | 12, 16, 4 | 16, 4 | 4 |
| Growth rate | 32, 16, 20, 40, 160 | 32, 32, 20, 40, 160 | 32, 32, 32, 40, 160 | 32, 32, 32, 32, 160 |
| GPU Time (ms) | 0.38 | 0.39 | 0.42 | 0.43 |
| Error (%) | 14.73 | 14.31 | 13.66 | 14.68 |
| #Params (M) | 13.64 | 13.92 | 14.37 | 11.47 |
| MAdd (G) | 6.27 | 6.72 | 6.90 | 5.76 |
| Flops (G) | 3.14 | 3.37 | 3.46 | 2.89 |

Because layer $l$ has $k$ (initial growth rate) feature maps, the number of channels is set to $k \times m^n$, where $n$ is referred to the largest value of $l$ divisible by $2^n$. Multiplier $m$ is used as the low dimension compression factor, and we set $m$ to 1.7 to compress the input channels into fewer output channels.

### E. Implementation Details

ThreshNet uses two 3 × 3 convolutions as the initial convolution layer, and uses one 1 × 1 convolution and 2 × 2 averaging pooling as the transition layer between the two consecutive Thresh-Blocks. At the end of the last Thresh-Block, the global average pooling is performed, and finally a softmax classifier is attached.

ThreshNet follows the global dense connection strategy employed by DenseNet and HarDNet, and all inputs to the Thresh-Block are used as part of its output. We propose three ThreshNet model architectures as ThreshNet28, ThreshNet79, and ThreshNet95. The detailed parameters are listed in TABLE I. When the input image size is 224 × 224, the feature image is reduced by 5 times, and the characteristic image sizes in the five Thresh-Blocks are 112 × 112, 56 × 56, 28 × 28, 14 × 14 and 7 × 7, respectively. For the reduction rate of excessive layer usage, we use 0.5 and 0.85, respectively, for dense connection and harmonic dense connection.

HarDNet completely uses harmonic dense connections, while ThreshNet uses dense connections in the first half of the network and uses harmonic dense connections in the last half of the network. Under this situation, there are more connections between layers than that of HarDNet, but the size of the network architecture is smaller than that of HarDNet. The numbers of the input channels used in HarDNet are 128, 256, 320, 640, and 1024, however the numbers of the input channels used in the proposed architecture are only 128, 192, 288, 480, and 960. Generally, in the network, the numbers and sizes of the feature map would increase upon the following layers. The sizes of the last several layers may be large and take a lot of hardware costs. Reducing the number of channels directly reduces the number of feature maps. According to the numbers of the input channels described earlier, the architecture of the proposed connection method is denser than that used by HarDNet. In addition, the correct rate of the proposed approach would be higher than that of HarDNet.

## IV. EXPERIMENTS

### A. Ablation Study

When introducing the threshold mechanism in Section III.A., we stated that setting the threshold value to the median or average number of input channels in all layers may work best. This conclusion is obtained through the ablation experiments. Taking ThreshNet79 as an example, the numbers of the input channels of the five blocks in the network are set to 128, 192, 288, 480, and 960, respectively. We designed ThreshNet79-a, ThreshNet79-b, ThreshNet79-c, and ThreshNet79-d, and set the threshold mechanism to execute before Blocks 2, 3, 4, and 5 respectively. Four network architectures use NVIDIA GPU RTX 3050 to evaluate the performance on CIFAR-10. The experimental results are listed in TABLE II. Setting the threshold value to 320 and executing the threshold mechanism before Block 4 can make the model perform the best. ThreshNet79-c has the highest accuracy with a similar inference time per image on CIFAR-10.

### B. Datasets

#### 1) CIFAR

The CIFAR-10 [28] dataset consists of 60,000 32 × 32 color images in 10 categories, with 6,000 images in each category. There are 50,000 training images and 10,000 test images.

#### 2) SVHN

The SVHN (street view house numbers) [29] is a dataset that can identify Arabic numerals in images from the real-world house numbers. Each image contains a set of "0-9" Arabic numerals. The training and testing sets contain 73,257 digits and 26,032 digits, respectively.

TABLE III
CIFAR-10 CLASSIFICATION RESULTS AND MODEL ARCHITECTURE PARAMETERS

| | GPU Time (ms) | Raspberry Pi Time (ms) | Error (%) | Flops (G) | MAdd (G) | Memory (MB) | #Params (M) | MenR+W (MB) |
|---|---|---|---|---|---|---|---|---|
| **ThreshNet 28** | **0.35** | **45.3** | **14.75** | **2.28** | **4.55** | **83.26** | **10.18** | **221.05** |
| ShuffleNet | 0.51 | 48.5 | 14.69 | 2.22 | 4.31 | 617.00 | 1.01 | 1009.03 |
| **ThreshNet 79** | **0.42** | **61.7** | **13.66** | **3.46** | **6.90** | **109.68** | **14.31** | **296.33** |
| HarDNet 68 | 0.44 | 62.4 | 14.66 | 4.26 | 8.51 | 49.28 | 17.57 | 181.97 |
| MobileNetV2 | 0.46 | 67.4 | 14.06 | 2.42 | 4.75 | 384.78 | 2.37 | 755.07 |
| SqueezeNet | 0.36 | — | 14.25 | 2.69 | 5.32 | 211.42 | 0.78 | 422.46 |
| ShuffleNetV2 | 0.37 | — | 18.07 | 2.23 | 4.39 | 280.29 | 1.37 | 535.97 |
| MobileNet | 0.38 | — | 16.12 | 2.34 | 4.63 | 230.84 | 3.32 | 474.13 |
| GhostNet | 0.45 | — | 19.96 | 0.15 | 0.29 | 40.05 | 5.18 | 98.84 |
| ResNeXt 50-32x4d | 0.45 | — | 16.96 | 4.12 | 8.22 | 109.69 | 25.55 | 317.43 |
| **ThreshNet 95** | **0.46** | — | **13.31** | **4.07** | **8.12** | **132.34** | **16.19** | **356.66** |
| Wide_ResNet 50-2 | 0.48 | — | 17.31 | 11.43 | 22.85 | 134.76 | 6.88 | 532.85 |
| HarDNet 85 | 0.50 | — | 13.89 | 9.10 | 18.18 | 74.65 | 36.67 | 313.42 |
| EfficientNet B0 | 0.52 | — | 13.40 | 1.51 | 2.99 | 203.74 | 3.60 | 421.54 |

The results for our network are in **boldface**. GPU Time is the inference time per image on NVIDIA RTX 3050, and Raspberry Pi Time is the inference time per image on Raspberry Pi 4.

## C. Training

For the sake of fairness, we did not use pre-training, fine-tuning, or image augmentation to improve the accuracy of ThreshNet, and used common hyperparameters to train the model for the experimental network. After fine-tuning, the network model can be improved to obtain high accuracies; however, the purpose of this experiment is to compare the performance of ThreshNet and other lightweight networks on mobile devices. For CIFAR-10 and SVHN datasets, we trained 150 and 60 epochs respectively, with a learning rate of 0.001, Adam as the optimizer, and the training batch size and test batch size are both set to 100. Therefore, it is not advisable to directly compare the results of this experiment with the accuracy of the state-of-the-art (SOTA) model.

## D. Comparisons

We use the inference time per image and test error rate to evaluate our algorithm and compare it with several state-of-the-art networks, including lightweight networks ShuffleNet V1, V2 [2][3], MobileNet V1, V2 [4][5], SqueezeNet [9], GhostNet [10], EfficientNet [11], and ResNeXt [30], Wide-ResNet [31], which improved the network ResNet [22].

## E. Testing

The trained network model is evaluated for performance on GPU and mobile device; GPU is RTX 3050 4GB, and mobile device is Raspberry Pi 4 Model B 4GB. The GPU was evaluated using Python 3.8, torch version 1.10.0, and the Raspberry Pi was evaluated using python 3.9, torch version 1.11.0.

## F. Classification Results on Datasets

Compared with other lightweight networks, ThreshNet achieves better image classification results on both CIFAR-10 and SVHN datasets.

TABLE III presents the test results on CIFAR-10 dataset. Using GPU for testing, ThreshNet28 has the shortest inference time per image compared to SqueezeNet, ShuffleNetV2, and MobileNet. The error rate of ThreshNet28 is 14.75%, which is only slightly higher than that of SqueezeNet (14.25%). We consider that ThreshNet28 outperforms other lightweight networks. Compared with the deeper networks, ThreshNet95 has the lowest error rate of 13.31%, and the inference time is less than its original network HarDNet85, dropping from 0.50ms to 0.46ms. This demonstrates that the performance of ThreshNet95 is superior in the similar scaled networks, and completely surpasses the original HarDNet85. Tested with Raspberry Pi, ThreshNet28 reduces inference time per image from 48.5ms to 45.3ms compared to ShuffleNet. Compared with the original network HarDNet68, ThreshNet79 reduces the error rate by 7%, and the inference time per image also drops from 62.4ms to 61.7ms. On mobile devices, reducing inference time by 1ms per image can reduce a lot of resource cost. The above results prove that ThreshNet can respond faster than MobileNetV2 and HarDNet on mobile devices such as Raspberry Pi.

In addition to the performance evaluation on CIFAR-10, we also evaluate the image classification on SVHN, and the test results are shown in TABLE IV. From the SVHN dataset we obtain conclusions similar to those on CIFAR-10. It is worth n-

TABLE IV
SVHN CLASSIFICATION RESULTS AND MODEL ARCHITECTURE PARAMETERS

|  | GPU Time (ms) | Raspberry Pi Time (ms) | Error (%) | Flops (G) | MAdd (G) | Memory (MB) | #Params (M) | MenR+W (MB) |
|---|---|---|---|---|---|---|---|---|
| **ThreshNet 28** | **0.36** | **41.5** | **6.34** | **2.28** | **4.55** | **83.26** | **10.18** | **221.05** |
| ShuffleNet | 0.50 | 45.6 | 6.34 | 2.22 | 4.31 | 617.00 | 1.01 | 1009.03 |
| MobileNetV2 | 0.46 | 52.1 | 6.68 | 2.42 | 4.75 | 384.78 | 2.37 | 755.07 |
| **ThreshNet 79** | **0.47** | **61.4** | **5.68** | **3.46** | **6.90** | **109.68** | **14.31** | **296.33** |
| HarDNet 68 | 0.48 | 64.6 | 7.03 | 4.26 | 8.51 | 49.28 | 17.57 | 181.97 |
| MobileNet | 0.36 | — | 7.36 | 2.34 | 4.63 | 230.84 | 3.32 | 474.13 |
| SqueezeNet | 0.37 | — | 6.13 | 2.69 | 5.32 | 211.42 | 0.78 | 422.46 |
| ShuffleNetv2 | 0.37 | — | 7.04 | 2.23 | 4.39 | 280.29 | 1.37 | 535.97 |
| GhostNet | 0.48 | — | 8.36 | 0.15 | 0.29 | 40.05 | 5.18 | 98.84 |
| EfficientNet B0 | 0.51 | — | 6.12 | 1.51 | 2.99 | 203.74 | 3.60 | 421.54 |
| **ThreshNet 95** | **0.53** | — | **6.89** | **4.07** | **8.12** | **132.34** | **16.19** | **356.66** |
| ResNeXt 50-32x4d | 0.55 | — | 7.73 | 4.12 | 8.22 | 109.69 | 25.55 | 317.43 |
| HarDNet 85 | 0.57 | — | 6.91 | 9.10 | 18.18 | 74.65 | 36.67 | 313.42 |
| Wide_ResNet 50-2 | 0.61 | — | 6.74 | 11.43 | 22.85 | 134.76 | 6.88 | 532.85 |

The results for our network are in **boldface**. GPU Time is the inference time per image on NVIDIA RTX 3050, and Raspberry Pi Time is the inference time per image on Raspberry Pi 4.

oting that the error rate of ThreshNet28 is the same as that of ShuffleNet (6.34%), but its inference time per image is shortened by 9%. This result once again proves that ThreshNet is more suitable for use on mobile devices, such as Raspberry Pi.

*G. Parameter Efficiency and Computation Efficiency*

TABLEs III and IV show multiplications and accumulations (MAdd), the total number of network parameters (Params), theoretical number of floating-point operations (FLOPs), and the sum of memory read and memory write time (MemR+W). ThreshNet performs model compression on HarDNet. Compared with HarDNet68, ThreshNet79 reduces the number of model parameters from 17.57M to 14.31M. Compared with HarDNet85, ThreshNet95 reduces the number of model parameters by 55%. Moreover, our model compression does not reduce the accuracy of the network model. Different versions of ThreshNet have higher accuracy rates than that of HarDNet.

The amount of model computation can be visualized by the inference time per image. As mentioned in the previous subsection, the amount of computation in ThreshNet is lower than that in similar scaled networks. This proves that our proposed approach for model compression and acceleration is feasible.

V. DISCUSSION

Similar to HarDNet, ThreshNet is also a CNN backbone network that is suitable for applications in different scenarios, such as SSD [32] and YOLO V4 [33] architectures for object detection scenes, U-Net architecture [34] for image semantic segmentation scenes, and FairMO architecture [35] for object tracking. Our experiments are also performed on Raspberry Pi for image classification, demonstrating the excellent model performance of ThreshNet. In fact, in addition to Raspberry Pi, there are mobile devices such as mobile phones and Jetson Nano. To better apply ThreshNet in real life, we will perform dense prediction tasks on more mobile devices in the future.

VI. CONCLUSION

This article proposes a new model compression and acceleration method, and proposes the network architecture of ThreshNet. ThreshNet is a lightweight network that can operate on mobile devices. Compared to other SOTA networks, our network has a shorter inference time and lower error rate, which is an important breakthrough on mobile devices. Because mobile devices are limited in terms of computing and battery power, the networks that run on them must have small model sizes and low computational complexity. ThreshNet complies with these requirements, and a lower computational load enables a lower inference time per image, thereby reducing the battery consumption.